\documentclass{article}

\usepackage{arxiv}

\usepackage[utf8]{inputenc} 
\usepackage[T1]{fontenc}    
\usepackage{hyperref}       
\usepackage{url}            
\usepackage{booktabs}       
\usepackage{amsfonts}       
\usepackage{nicefrac}       
\usepackage{microtype}      
\usepackage{lipsum}
\usepackage{graphicx}
\graphicspath{ {./images/} }

\usepackage{algorithm,algorithmic}%
\usepackage{multirow}
\usepackage{amsmath}

\DeclareMathOperator\softmax{softmax}

\title{Fusion-GRU: A Deep Learning Model for Future Bounding Box Prediction of Traffic Agents in Risky Driving Videos }

\author{
 Muhammad Monjurul Karim \\
  Department of Civil and Environmental Engineering\\
  University of Washington\\
  Seattle, WA 98125, USA \\
  \texttt{mmkarim@uw.edu} \\
   \And
 Ruwen Qin \\
  Department of Civil Engineering\\
  Stony Brook University\\
  Stony Brook, NY 11794, USA \\
  \texttt{ruwen.qin@stonybrook.edu} \\
  \And
 Yinhai Wang \\
  Department of Civil and Environmental Engineering\\
  University of Washington\\
  Seattle, WA 98125, USA \\
  \texttt{yinhai@uw.edu} \\
}

\begin{document}
\maketitle
\begin{abstract}
To ensure the safe and efficient navigation of autonomous vehicles and advanced driving assistance systems in complex traffic scenarios, predicting the future bounding boxes of surrounding traffic agents is crucial. However, simultaneously predicting the future location and scale of target traffic agents from the egocentric view poses challenges due to the vehicle's egomotion causing considerable field-of-view changes. Moreover, in anomalous or risky situations, tracking loss or abrupt motion changes limit the available observation time, requiring learning of cues within a short time window. Existing methods typically use a simple concatenation operation to combine different cues, overlooking their dynamics over time. To address this, this paper introduces the Fusion-Gated Recurrent Unit (Fusion-GRU) network, a novel encoder-decoder architecture for future bounding box localization. Unlike traditional GRUs, Fusion-GRU accounts for mutual and complex interactions among input features. Moreover, an intermediary estimator coupled with a self-attention aggregation layer is also introduced to learn sequential dependencies for long range prediction. Finally, a GRU decoder is employed to predict the future bounding boxes. The proposed method is evaluated on two publicly available datasets, ROL and HEV-I. The experimental results showcase the promising performance of the Fusion-GRU, demonstrating its effectiveness in predicting future bounding boxes of traffic agents.
\end{abstract}

\section{Introduction}
In recent years, the emergence of autonomous driving has attracted significant attention and achieved remarkable milestones \cite{eskandarian2019research, muhammad2020deep}. While autonomous driving offers convenience to people and addresses industry's evolving needs, it also raises concerns about traffic safety. Notably, from 2014 to July 25, 2023, 627 autonomous vehicle collisions were reported in California, USA \cite{ca_dmv_2021}. Moreover, the World Health Organization's 2018 global status report on road safety reveals a distressing figure of 1.35 million annual fatalities due to traffic crashes worldwide \cite{world2018global}. Most of these crashes are caused by human error \cite{eskandarian2019research}, which highlights the need for technologies that can assist drivers and improve safety on road, especially in mixed traffic conditions where most vehicles are operated by humans. Thus, anticipating future locations of traffic agents becomes a critical safety-enhancement feature for autonomous vehicles and countless human-driven vehicles. Accurately predicting the movement and future locations of different traffic agents, including pedestrians and vehicles, is essential for the secure navigation of autonomous vehicles and advanced driving assistance systems. By successfully anticipating the future location and scale of these traffic agents, a vehicle can plan secure and socially-aware paths, thus proactively avoiding crashes or near misses \cite{liang2019peeking}. 

Early research focused primarily on predicting the future location of individual agents \cite{liang2019peeking, gupta2018social}. However, simultaneously predicting future bounding boxes of multiple agents presents considerable challenges. Recurrent neural networks (RNNs), particularly long short-term memory networks (LSTMs) and gated recurrent units (GRUs), have exhibited promise in this direction \cite{yao2019unsupervised,wang2022stepwise}. Nevertheless, most RNN-based models suffer from performance degradation over time since they rely on recurrently predicting future bounding boxes based on previous outputs. Very recent works uses Transformer based methods for future location estimation of traffic agents \cite{yu2020spatio, li2022sit, yao2022end}. Despite advancements in deep learning based methods for future object localization in normal driving scenarios, this problem in the risky driving scenarios remains largely unsolved. The complexity arises from dealing with risky or anomalous events, wherein traffic agents demonstrate sudden or abrupt motion changes. Furthermore, capturing the inter-dependencies between the outputs of successive frames presents an additional challenge. 

To tackle these challenges, this paper proposes an encoder-decoder based architecture for future traffic agent localization. The proposed framework leverages multiple sources of information, including location-scale data, monocular depth information, and optical flow data for effective representation learning. To achieve this, we introduced a novel Fusion-Gated Recurrent Unit (Fusion-GRU) encoder, specifically designed to capture complex interactions among information cues and transform them into hidden representations. This encoding process effectively learns the complex relationships between crucial cues within temporal sequences, leading to enhanced future traffic agent localization performance, particularly in cluttered risky driving scenes, when the available observation length is short. To further enhance the prediction performance for long-range scenarios, an intermediary estimator and a self-attention aggregation layer are introduced. The intermediary estimator generates intermediary bounding boxes, facilitating the learning of sequential dependencies in predicting the output of future frames. Additionally, the self-attention aggregation layer plays a crucial role in adaptively learning the importance of different future frames and reducing error accumulation in the recurrent unit. The major contributions of this paper are summarized below:

\begin{itemize}
    \item Development of the novel Fusion-GRU encoder, effectively encoding complex interactions among input information cues into hidden representations.  
    \item Extraction and fusion of scene- and traffic agent-specific spatio-temporal flow features, enabling better capture of motion information from traffic agents and the overall scene.
    \item Introduction of an intermediary estimator in conjunction with a self-attention aggregation layer to generate intermediary bounding boxes, facilitating the learning of sequential dependencies for long-term predictions.   

\end{itemize}

The remainder of this paper is organized as follows. Section 2 provides a summary of the literature review. Section 3 delineates the methodology of our proposed framework. Section 4 describes the implementation details and presents experimental evaluation of the proposed method, including qualitative and quantitative analysis of the results. Finally, Section 5 presents the conclusion and future work. 

\section{Literature Review} 
The problem of future object localization has been extensively studied. Early works in this field aimed to address this problem using static surveillance cameras from a bird's eye view perspective. Some of the early methods include recurrent neural networks (RNNs). For example, long short-term memory networks (LSTMs) and gated recurrent units (GRUs), have been employed in an encoder-decoder (ED) format to encode past observations and decode future locations \cite{choi2019looking, sadeghian2018car}. Additional inputs, such as heterogeneous information on neighbors, maps, and semantic actions, have been utilized to improve the accuracy of future location prediction. Notable among these methods is the work by Alahi et al.\cite{alahi2016social}, who proposed a Social-LSTM to model pedestrian trajectories and their interactions with neighbors. The social pooling module of their approach was further improved by \cite{gupta2018social} to capture global context. SoPhie \cite{sadeghian2019sophie} applies generative models to account for the uncertainty in future paths, while Lee et al. \cite{lee2017desire} use RNNs with conditional variational autoencoders (CVAEs) to generate multi-modal predictions. AEE-GAN \cite{lai2020trajectory} use attention mechanisms to address complex scenes with diverse interacting agents. Other works \cite{ivanovic2019trajectron,tang2019multiple, kosaraju2019social} have also introduced graph-based recurrent models that simultaneously predict potential trajectories of multiple objects. Recent works have adopted Transformer-based methods \cite{yu2020spatio, li2022sit, yao2022end} for future location prediction, leveraging their ability to capture long-range dependencies and complex interactions among agents in the datasets.

However, while the above-mentioned methods are designed for third-person views from static cameras, they have limitations in scenarios involving ego motion, such as driving scenarios. Consequently, recent works have focused on jointly modeling the motion of observed objects and the ego-camera. Bhattacharyya et al. \cite{bhattacharyya2018long} propose Bayesian LSTMs to model observation uncertainty and predict the distribution of future locations. Yagi et al. \cite{yagi2018future} incorporate multi-modal data, such as human pose, scale, and ego-motion, as cues in a convolution-deconvolution (Conv1D) framework to predict future pedestrian locations. Yao et al. \cite{yao2019egocentric} introduce a multi-stream encoder-decoder that separately captures both object location and appearance. Makansi et al. \cite{makansi2020multimodal} estimate a reachability prior for objects from the semantic map and propagate them into the future. More recently, Wang et al. \cite{wang2022stepwise} have developed a stepwise goal-driven network that predicts a stepwise goal for each timestep to guide the training methods and improve the prediction. Another notable work is PedFormer \cite{rasouli2023pedformer}. This method employs a transformer encoder to effectively encode crucial elements such as appearance, ego-vehicle motion, and previous trajectory information. Additionally, this method utilizes an attention module to capture and encode the interactions between the encoded information, further enhancing the model's performance.

While these methods have laid a strong foundation in traffic agent's future bounding box localization, it is important to note that most of them do not explicitly model and learn the intrinsic relations among different cues, such as motion, location, scale, and distance between different agents. These intrinsic relations and mutual interactions play a crucial role in reasoning and accurately predicting the future bounding box of traffic agents. Additionally, these methods overlook the significance of incorporating correlations between the outputs of different frames into the learning process of deep neural networks.

\section{Methodology}
Our objective is to forecast the bounding boxes for all the traffic agents in the next $N$ frames, based on their observed bounding boxes in the current video frame with $M$ traffic agents. The observed bounding box of the $i$-th traffic agent at time step $t$ can be represented as $\pmb{b}_{t,i} = [x_{t,i}, y_{t,i}, w_{t,i}, h_{t,i}]$, where ($x_{t,i}, y_{t,i}$) denotes the center location, and $w_{t,i}$ and $h_{t,i}$ represent the width and height, respectively, of the bounding box for the $i$-th object in the current frame $t$. The output set of future bounding boxes for the $i$-th object can be described as $\pmb{B}_{t,i} = \{ \pmb{b}_{(t+1),i}, \dots , \pmb{b}_{(t+N),i} \}$, encompassing the predicted bounding boxes for the $i$-th object in the subsequent $N$ frames.

\begin{figure*}[htb]
\centering
\includegraphics[width=\textwidth]{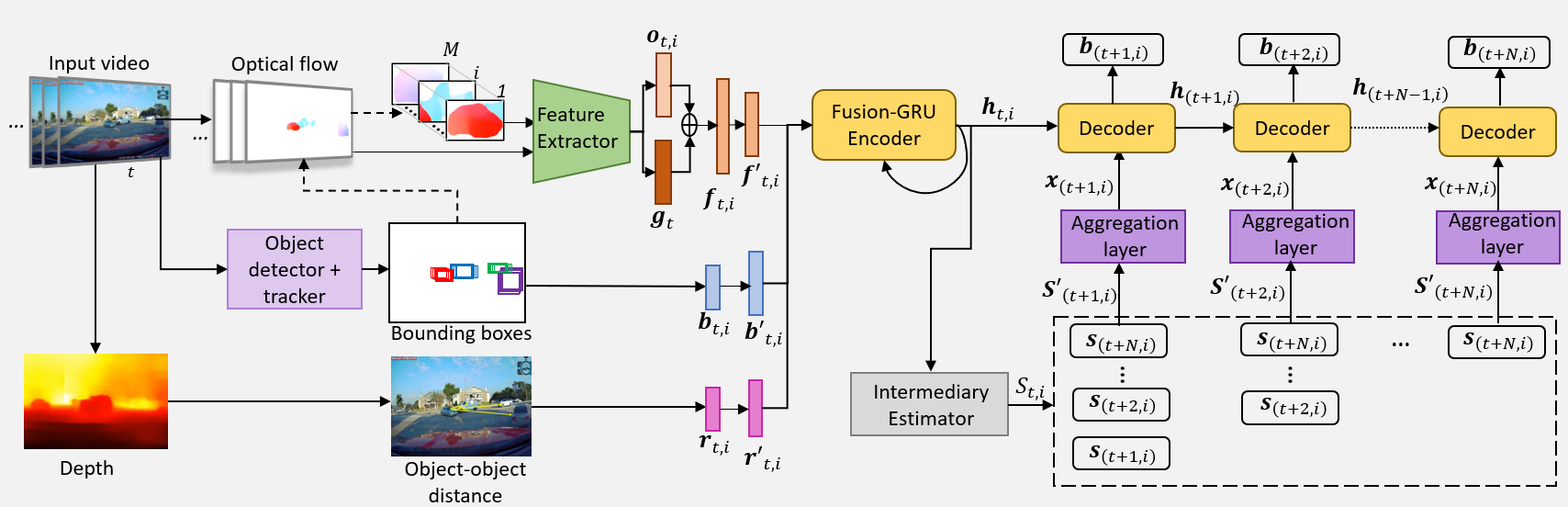}
\caption{Overview of the proposed Fusion-GRU based encoder-decoder framework for future bounding box prediction. }
\label{fig_overview}
\end{figure*}

FIGURE \ref{fig_overview} illustrates our proposed framework. At first, the network takes the input video sequence frame by frame. A depth estimator is used to get dense depth map, which is then used to approximate distances between traffic agents. An object detector detects the bounding boxes of the traffic agent, while a tracker tracks them. An optical flow estimator calculates the optical flow. Based on the bounding box information, we crop the traffic agents from the flow images. Then, a feature extractor extracts the CNN features from both the cropped objects and the entire frame. These two features are concatenated and fed to the Fusion-GRU encoder along with bounding box feature and distance feature vector to encode the information into hidden representation. Using the hidden representation, the intermediary estimator produces a set of intermediary bounding boxes. Then, a self-supervised aggregation layer converts them into a single feature vector. Taking that feature vector along with the previous hidden representation a GRU decoder generates the bounding boxes for the future frames. Detailed description of the proposed method is below.

\subsection{Feature Extraction}
Our proposed method uses a pretrained object detector YOLOv5 \cite{ge2021yolox} to detect traffic agents with their corresponding bounding boxes in each video frame. The number of detected traffic agents, denoted as $M$, may vary from frame to frame. To establish object associations across different video frames, we employ the multi-object tracker DeepSort \cite{wojke2017simple}. 

The variation in pixel-level motion across video frames serves as a critical cue for identifying objects with unusual movement patterns. Therefore, this study uses a pre-trained RAFT model \cite{teed2020raft} to extract the optical flow image for each frame. Then, the bounding box regions detected from the flow images are cropped, yielding $M$ object-level flow images for each detected traffic agent. A ResNet50 \cite{he2016deep} feature extractor turns the frame-level flow image into a global feature vector, $\pmb{g}_t(\in\mathbb{R}^{D})$. Similarly, the feature extractor converts each of the object-level flow image into another feature vector, $\pmb{o}_{t,i}(\in\mathbb{R}^{D})$. Here, $D$ is the dimension of these feature vectors. Then, $\pmb{o}_{t,i}$ is concatenated with $\pmb{g}_{t}$ to create the overall flow feature vector for the $i$th object, $\pmb{f}_{t,i}(\in\mathbb{R}^{2D})$:

\begin{equation}
    \pmb{f}_{t,i}=[\pmb{o}_{t,i}; \pmb{g}_{t}],
\end{equation}
which captures the object's motion feature in the driving scene. Then, with a fully connected layer $\phi$, the flow feature vector $\pmb{f}_{t,i}$ is converted into a lower dimensional vector,  $\pmb{f}'_{t,i}\in(\mathbb{R}^{2d})$:

\begin{equation}
    \pmb{f}'_{t,i} = \phi(\pmb{f}_{t,i};\pmb{\theta}_{0}),
\label{eq:flow}
\end{equation}
where $\pmb{\theta}_0$ are learnable parameters of the fully connected layer.

Spatial dynamics of traffic agents over time can be captured by observing changes in the location and scale of their bounding boxes in successive frames. Bounding box $\pmb{b}_{t,i}$ is then passed through a fully connected layer, transforming it into a feature vector in a $d$ dimensional space, $\pmb{b}'_{t,i}(\in\mathbb{R}^d)$.

Monocular depth is estimated using DenseDepth model \cite{alhashim2018high},  which was pre-trained on the KITTI \cite{geiger2013vision} dataset. Afterward, we converted the depth map values to actual distance measurements in meters, using a scale factor of 80 meters as provided by \cite{alhashim2018high}. A distance vector $\pmb{r}_{t,i}$ is created by taking the distance of the closest 6 traffic agents from the traffic agent $i$. This vector is then upsampled using a fully connected layer, yielding the distance feature vector $\pmb{r}'_{t,i} \in \mathbb{R}^k$. Here, $k$ represents the dimension of the feature vector.

\begin{figure*}[htb]
\centering
\includegraphics[width=\textwidth]{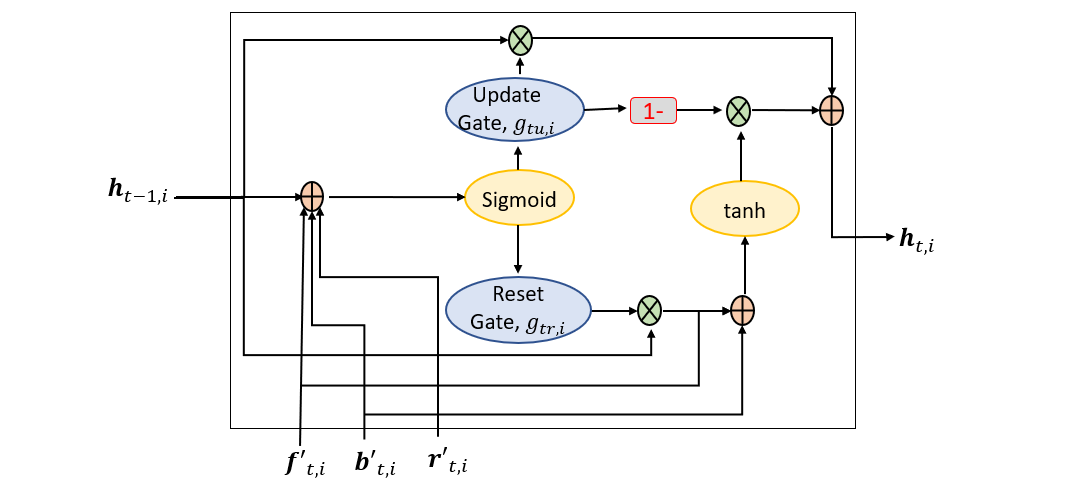}
\caption{Detailed structure of Fusion-GRU.}
\label{fig_fusionGRU}
\end{figure*}

\subsection{Fusion-GRU Encoder}
Extracted features mentioned in previous section are all highly correlated with each other for the prediction of the traffic agent's future bounding box. To leverage the intrinsic interactions among these features, this paper proposes to integrate all the features into the learning of recurrent unit. Consequently, we present a modified GRU named Fusion-GRU, designed to encode this information into a hidden representation and capture the underlying relationships. FIGURE \ref{fig_fusionGRU} shows the structure of the proposed Fusion-GRU.

In the proposed Fusion-GRU model, the data flow is described by a set of equations (\ref{eq:resetgate_5}-\ref{eq:ht_5}). First, the reset gate $(\pmb{g}_{tr,i})$ for each traffic agent \(i\) at time step \(t\) is computed by applying a sigmoid activation function by combining flow feature (\(\pmb{f}'_{t,i}\)), bounding box feature (\(\pmb{b}'_{t,i}\)), distance feature (\(\pmb{r}'_{t,i}\)), and the previous hidden state (\(\pmb{h}_{(t-1),i}\)), each linearly transformed by their respective weight matrices $\pmb{W}$'s. Subsequently, the reset gate is used to compute \(\pmb{e}_{t,i}\) for $i$-th traffic agent at time step \(t\) by incorporating the flow feature, bounding box feature, and distance feature representations with the previous hidden state, using additional learnable weight matrices ($\pmb{W}$'s) and the hyperbolic tangent activation function. The update gate (\(\pmb{g}_{tu,i}\)) is then computed similar to the reset gate calculation. Finally, the updated hidden state (\(\pmb{h}_{t,i}\)) for each traffic agent \(i\) at time step \(t\) is obtained by blending the reset gate-controlled candidate hidden state with the previous hidden state, influenced by the update gate.

\begin{equation}  
\pmb{g}_{tr,i} = \sigma(\pmb{W}_{fr}\pmb{f}'_{t,i} + \pmb{W}_{br}\pmb{b}'_{t,i}+ \pmb{W}_{dr}\pmb{r}'_{t,i} + \pmb{W}_{hr}\pmb{h}_{(t-1),i}),
\label{eq:resetgate_5}
\end{equation} 

\begin{equation}
\pmb{e}_{t,i} = \tanh (\pmb{g}_{tr} \odot (\pmb{h}_{t-1}) + \pmb{W_{f1}}\pmb{f}'_{t,i} + \pmb{W_{b1}}\pmb{b}'_{t,i} +\pmb{W_{d1}}\pmb{r}'_{t,i} ), 
\end{equation}

\begin{equation} 
\pmb{g}_{tu,i} = \sigma(\pmb{W}_{fu}\pmb{f}'_{t,i} + \pmb{W}_{bu}\pmb{b}'_{t,i}+ \pmb{W}_{du}\pmb{r}'_{t,i} + \pmb{W}_{hu}\pmb{h}_{(t-1),i}), 
\end{equation} 

\begin{equation} 
    \pmb{h}_{t,i} = (1- \pmb{g}_{tu,i}) \odot \pmb{e}_{t,i}  + \pmb{g}_{tu,i} \odot \pmb{h}_{(t-1),i}, 
\label{eq:ht_5}
\end{equation} 
Here, $\sigma$ represents the sigmoid activation, $\odot$ is the element-wise product operator.

\subsection{Intermediary Estimator}
The dependencies between different frames are not equal. There are some relationships involved between the objects bounding boxes in different video frames. It is important to capture these intrinsic sequential relationships between the predicted future bounding boxes across frames. To capture these dependencies, we introduce an intermediary estimator. This estimator serves to predict intermediate future bounding boxes before reaching the final decoder. By leveraging these predicted intermediary bounding boxes, the training process can be guided more effectively.
This module can be defined as:

\begin{equation} 
    {\pmb{S}_{t,i} = \gamma[\phi(\phi(\pmb{h}_{t,i};\pmb{\theta_0});\pmb{\theta}_1)]} = \{\pmb{s}_{(t+1,i)}, \pmb{s}_{(t+2,i)}, \dots ,\pmb{s}_{(t+N,i)} \}, 
\label{eq:It}
\end{equation}

Where, $\pmb{S}_{t,i}$ is the set of intermediary bounding boxes for the next $N$ frames, which is generated by passing the hidden representation $\pmb{h}_{t,i}$ from the Fusion-GRU through two fully connected layers ($\phi$). $\pmb{\theta}$'s are the learnable parameters in the fully connected layers. 


\begin{algorithm}[!ht]
\caption{Subset Formation Procedure}\label{algorithm_1}
\begin{algorithmic}
\STATE {$\pmb{S}_{t,i} \gets \{\pmb{s}_{(t+1,i)}, \pmb{s}_{(t+2,i)} \dots \pmb{s}_{(t+N,i)}\}$}; \COMMENT{$\pmb{S}_{t,i}$ is the set of intermediary bounding boxes}

\FOR{$j \gets 1\ to \ N$} 
\STATE {$\pmb{S}'_{(t+j,i)} \gets \{ \pmb{S}_{t,i}[j:N] \} \gets \{\pmb{s}_{(t+j,i)}: \pmb{s}_{(t+N,i)}\} $}; \COMMENT{$j$ is the index for future timestep}
\ENDFOR
\end{algorithmic}
\end{algorithm}

For each future timestep $j$, a subset $\pmb{S}'_{(t+j,i)}$ is extracted from the set $\pmb{S}_{t,i}$ using Algorithm \ref{algorithm_1}. These subsets are created to fed to the decoder to mitigate the influence of older time steps and emphasize on the recent time steps.

\subsection{Self-attention Aggregation Layer}
The proposed method employs a self-attention aggregation layer for each subset of intermediary bounding boxes. This self-attention mechanism helps to focus on the most relevant information and create a consolidated representation. Moreover, this aggregation layer helps to minimize the error accumulation within the recurrent unit.

The data flowing through the self-attention aggregation layer are expressed mathematically in equations (\ref{eq:zt}-\ref{eq:xt}):

\begin{equation}
\pmb{z}_{(t+j,i)} = ReLU(\phi(\pmb{S}'_{(t+j,i)} ;\pmb{\theta}_z) )
\label{eq:zt}
\end{equation}

\begin{equation}
\pmb{W}_{saa} = \softmax (\pmb{W}_{sa} \pmb{z}_{(t+j,i)}))
\label{eq:wsaa}
\end{equation}

\begin{equation}
\pmb{x}_{(t+j,i)} = \sum_{j=j}^N (\pmb{W}_{saa} \odot \pmb{z}_{(t+j,i)})
\label{eq:xt}
\end{equation}
wherein, $\pmb{W}_{sa}$ and $\pmb{\theta}_z$ are learnable parameters, $\phi$ is the fully connected layer, and $\odot$ indicates the element-wise product operator.


\subsection{GRU Decoder}
Finally, a GRU decoder takes the aggregated feature $\pmb{x}_{(t+j,i)}$ to decode this feature information into hidden representation $\pmb{h}_{(t+j,i)}$:

\begin{equation}
\pmb{h}_{(t+j,i)} = GRU(\pmb{x}_{(t+j,i)}, \pmb{h}_{(t+j-1,i)}; \pmb{\theta}_h)
\label{eq:ht_decoder}
\end{equation}

where, $\pmb{\theta}_h$ is the learnable parameters of the GRU.

Then, two fully connected layers generate the future bounding box $\pmb{b}_{(t+j,i)}$ for the $j$-th frame.

\begin{equation}
\pmb{b}_{(t+j,i)} = \phi(\phi(\pmb{h}_{(t+j,i)} ;\pmb{\theta}_{b1}); \pmb{\theta}_{b2} )
\label{eq:pred}
\end{equation}

where, $\theta_b$'s are the learnable parameters of the fully connected layers.

\subsection{Loss Function}
The loss function for the proposed method consists of two terms. The first loss term is computed for the final output bounding boxes, which are the predicted bounding boxes for the target frames. The second loss term is computed for the intermediary bounding boxes, which are the predicted bounding boxes from the intermediary estimator. To calculate these losses, smooth L1 loss function is employed. The smooth L1 loss function $\mathcal{L}_{\mathrm{smoothL1}}(Y_t, \hat{Y_t}) $ can be defined as:
\begin{equation}
\mathcal{L}_{\mathrm{smoothL1}}(Y_t,\hat{Y_t}) =
\begin{cases}
0.5 \cdot (Y_t - \mathrm{\hat{Y}_t})^2, & \text{if}\ |Y_t - \mathrm{\hat{Y}_t}| < 1 \cr
|Y_t - \mathrm{\hat{Y}_t}| - 0.5, & \text{otherwise}
\end{cases}
\end{equation}

where, $Y_t$ is the prediction and $\hat{Y}_t$ is the ground truth. Therefore, the total loss for the $i$-th object at frame $t$ is computed as:

\begin{equation}
\label{eq_loss}
   \mathcal{L}_{\mathrm{t,i}} = \sum_{j=j}^N \mathcal{L}_{\mathrm{smoothL1}}(\pmb{b}_{(t+j,i)},\hat{\pmb{Y}}_{(t+j,i)}) + \lambda \mathcal{L}_{\mathrm{smoothL1}}({\pmb{s}_{(t+j,i)},\hat{\pmb{Y}}_{(t+j,i)})}
\end{equation}

In equation \ref{eq_loss}, the left loss term is computed using the final predicted bounding boxes and the right loss term is computed using intermediary bounding boxes. $\lambda$ is a hyper-parameter that helps to adjust the relative importance between the two losses. This study sets $\lambda$ to 0.3.

\section{Implementation and Experimental Evaluation}
The proposed method is built using PyTorch \cite{paszke2019pytorch}. Model training and testing are performed using an Nvidia Tesla V100 GPU with 32GB of memory. All the input frames are resized to 224$\times$224 before feeding to the ResNet50 \cite{he2016deep} feature extractor. Feature vectors $\pmb{g}_{t,i}$ and $\pmb{o}_{t,i}$ are obtained by applying an average pooling operation to the output of the ResNet50 feature extractor, resulting in a dimension of 2,048 ($D$) . A fully-connected layer further reduces the dimension to 256 ($d$). The dimension of the hidden representations is 256 ($d$) and distance vector is 32 ($k$). A learning rate of 0.001 is used to train the network, and ReduceLROnPlateau is used as the learning rate scheduler. Adam optimizer is used to optimize the network for 30 epochs and the best model is selected.

\subsection{The Dataset}
\subsubsection{Risky Object Localization (ROL) } Risky Object Localization (ROL) \cite{karim_am_net} dataset is a dashcam dataset contains 1000 crash videos with diverse environmental attributes. The dataset includes spatial, temporal, and categorical annotations. Each video clip is of 5 seconds length and encoded at 20 frames per second (fps) with a resolution of 1080 $\times$ 720. Different videos contain varying numbers of risky or risk-inducing traffic agents.

\subsubsection{Honda Egocentric View-Intersection (HEV-I)}
The Honda Egocentric View-Intersection (HEV-I) dataset \cite{yao2019egocentric} consists of 230 egocentric dashcam videos recorded in urban driving scenarios. The dataset portrays normal driving conditions without any anomalous events and specifically focuses on intersections. Each video ranges from 10 to 60 seconds in length, encoded at 10 frames per second (fps), and has a resolution of 1280 $\times$ 640.

\subsection{Evaluation Metrics}

\subsubsection{Evaluation of Location Prediction} 
To evaluate the performance of location or trajectory prediction, this paper uses Final Displacement Error (FDE) and Average Displacement Error (ADE) metrics. The FDE measures the Euclidean distance between the predicted final position and the true final position of the traffic agent. It is defined as follows:
\begin{equation}
\mathrm{FDE}_t = \left|{c}_{(t+N,i)} - {\hat{c}}_{(t+N,i)}\right|_2
\end{equation}

where, ${c}_{(t+N,i)}$ and ${\hat{c}}_{(t+N,i)}$ are the center locations of the predicted and ground truth bounding boxes of the object, respectively, at time $t+N$. Euclidean distance between the two positions is computed using the $L_2$ norm.

The ADE measures the average Euclidean distance between the predicted positions and the true positions of the traffic agent at each time step in the predicted trajectory. It is defined as follows:
\begin{equation}
\mathrm{ADE}_t = \frac{1}{N} \sum_{j=1}^{N} \left|{c}_{(t+j,i)} - {\hat{c}}_{(t+j,i)}\right|_2
\end{equation}

where, ${c}_{(t+j,i)}$ and ${\hat{c}}_{(t+j,i)}$ are the center locations of the predicted and ground truth bounding boxes of the $i$-th object, respectively, at time $t+j$.

\subsubsection{Evaluation of Bounding Box Prediction} 
To evaluate the performance of scale or bounding box prediction, Final Intersection over Union (FIoU) and Average Intersection over Union (AIoU) metrics are used. Intersection over Union (IoU) measures the overlap between the ground truth and the predicted bounding boxes. It takes the ratio between their intersection to their union. FIoU measures the IoU at the final position of the traffic agent across the prediction horizon. AIoU computes the average IoU of all the future predicted time steps of the traffic agent.

\subsection{Experimental Evaluation}
\subsubsection{Quantitative Result}
Performance of the proposed method is evaluated and compared with FOL-X \cite{yao2019unsupervised} and SGDNet  \cite{wang2022stepwise} on HEV-I dataset. The results are presented in TABLE-\ref{tab:trajectory_hevi}. While the performance of FOL-X and SGDNet was cited from  \cite{yao2019unsupervised, wang2022stepwise}, it is worth noting that SGDNet did not report their FDE$_{0.5}$ and FIoU$_{0.5}$ on HEV-I dataset.
\begin{table}[H]
\centering
\caption{{Comparison of the proposed method with other methods on HEV-I dataset}}

    \label{tab:trajectory_hevi}
\begin{tabular}{l|r|r|r|r}
\textbf{Methods} & \textbf{ADE$_{0.5}$ (pxl)} & \textbf{FDE$_{0.5}$ (pxl)} & \textbf{FIoU$_{0.5}$ (\%)} &\textbf{ADE$_{1.0}$ (pxl)} \\ \hline
FOL-X \cite{yao2019unsupervised}           & 6.7                & 11.0 & 85.0 & 12.6                \\
SGDNet  \cite{wang2022stepwise}          & 6.3               & -     & -       & 11.4      \\
\textbf{Ours}    & 5.5               & 8.3 & 85.2   & 11.2           \\ \hline
\end{tabular}
\end{table}

In the performance evaluation, we examined two prediction horizons: 0.5 second and 1 second. The results showed that our proposed method outperformed other methods. For the 0.5-second prediction horizon, the Fusion-GRU method achieved an ADE$_{0.5}$ of 5.5 pixels, which was significantly lower than the ADE$_{0.5}$s of 6.7 pixels for FOL-X and 6.3 pixels for SGDNet. Moreover, the proposed method exhibited the shortest FDE$_{0.5}$ of 8.3 pixels. Additionally, in terms of location and scale prediction accuracy, the Fusion-GRU method achieved the best performance, as indicated by the highest FIoU$_{0.5}$ of 85.2\% after 0.5 seconds. When extending the prediction horizon to 1 second, our proposed Fusion-GRU method maintained its superiority, achieving an ADE$_{1.0}$ of 11.2 pixels, while FOL-X and SGDNet obtained ADE$_{1.0}$s of 12.6 and 11.4 pixels, respectively. 

The proposed method also achieved very promising performance on the challenging ROL \cite{karim_am_net} dataset. Fusion-GRU achieved 15.3 pixel FDE$_{0.5}$, 9.0 pixel ADE$_{0.5}$, 62.9 \% FIoU$_{0.5}$, and 73.4\% AIoU$_{0.5}$. These results indicate that, despite the presence of situations in the ROL dataset where the ego vehicle or risky traffic agents undergo abrupt changes in their motion paths, the proposed method successfully predicts their future bounding boxes. This achievement highlights the effectiveness of the proposed approach in handling real-world driving situations with dynamic and unpredictable traffic behaviors.

\begin{figure*}[htb]
\centering
\includegraphics[width=\textwidth]{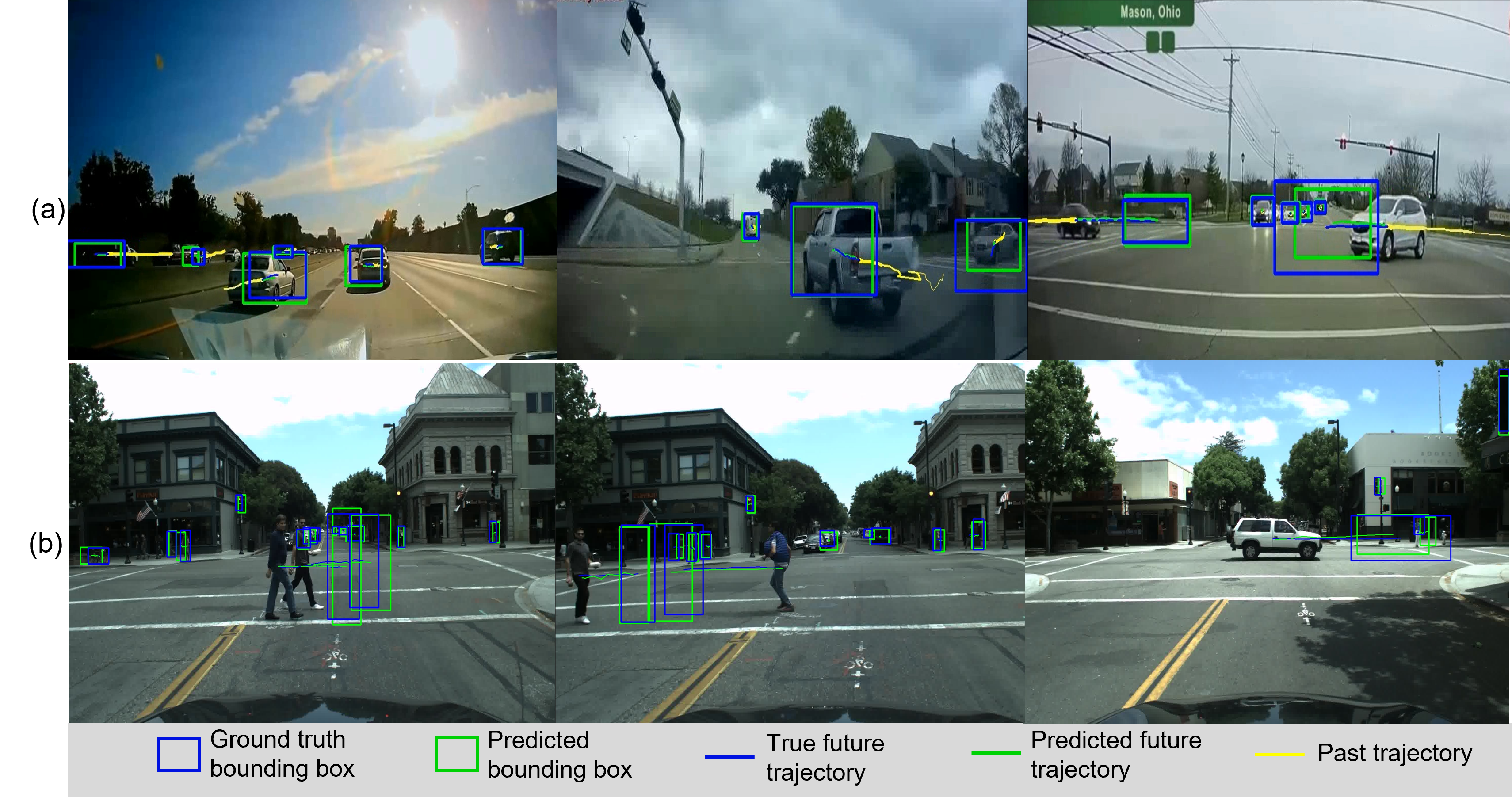}
\caption{Examples of future bounding box prediction in driving scene: (a) samples from ROL dataset with 0.5 second prediction horizon, (b) samples from HEV-I dataset with 1 second prediction horizon.}
\label{fig_example}
\end{figure*}

\subsubsection{Result Visualization}
FIGURE \ref{fig_example} illustrates some of the examples generated by the proposed method. Each example consists of a blue bounding box representing the ground truth and a green bounding box representing the predicted bounding box at the final time step of the prediction horizon. Additionally, blue and green curves depict the ground truth and predicted future trajectories, respectively. 

FIGURE \ref{fig_example} (a) shows examples from ROL dataset. These examples contain risky situations with traffic agents potentially heading towards a collision. The past trajectories of traffic agents are shown with a yellow curve to visualize their intricate trajectories. Complex scenarios with ego motion and multiple vehicles are present in these examples. The predicted trajectories and bounding boxes align well with the ground truth in the left two examples. The right example in Figure \ref{fig_example}(a) illustrates a scenario involving a white SUV turning left from the right and a black sedan moving from the left to the right. For this case, both the trajectory and bounding box predictions for the black car match closely with the ground truth. Though there is a slight discrepancy in the bounding box prediction for the white vehicle, the predicted trajectory aligns well with the ground truth.

FIGURE \ref{fig_example} (b) shows results from HEV-I dataset. In the left sample frame, two pedestrians are observed crossing the road at a walking pace. In the middle example, a pedestrian is running to cross the road. The right sample exhibits a vehicle moving from the left side to the right side of the road. Notably, in all these cases, the predicted trajectories and bounding boxes by the proposed method closely align with the ground truth. These examples indicate that the proposed method can successfully predict the future bounding box of different traffic agents across diverse scenarios.

\section{Conclusions}
This paper presented an encoder-decoder based future bounding box prediction method in driving videos. The proposed Fusion-GRU outperforms traditional GRU by effectively integrating input features to capture complex interactions among crucial cues from limited observation frames, enabling more accurate predictions of future bounding boxes. The incorporation of the intermediary estimator and self-supervised aggregation layer further enhances the sequential dependencies learning between predictions from different frames, providing a significant advancement in the performance of future bounding box prediction models for driving scenes. Experimental results demonstrate the superiority of the proposed Fusion-GRU method over existing approaches in accurately predicting future bounding boxes.

However, it is essential to acknowledge some limitations. The reliance on a forward-facing dashcam as the sole input sensor may have drawbacks such as low field of view or vision obstruction in challenging environmental conditions. To address these limitations, future research should explore sensor fusion as a solution, combining different sensor types to mitigate individual sensor drawbacks and enhance prediction robustness. Advancing and harmonizing sensor technologies will play a pivotal role in achieving full driving automation.

The integration of Vehicle-to-Vehicle (V2V) communication holds tremendous potential in leveraging the future traffic agent localization of risky traffic agents. By sharing this information among vehicles, we can enhance overall road safety and traffic efficiency. As V2V communication becomes more prevalent, the impact of dashcam-based future bounding box anticipation will be further amplified.

Furthermore, future work could focus on incorporating scene context to capture interactions between traffic agents and their surroundings or infrastructure. By considering these interactions, the performance of future bounding box prediction models can be further improved, leading to even more accurate and context-aware predictions.


\bibliographystyle{unsrt}  


\bibliography{references}

\end{document}